%% file: neurips_2022.tex
\documentclass{article}

\usepackage[preprint]{neurips_2022}

\usepackage[utf8]{inputenc} 
\usepackage[T1]{fontenc}    
\usepackage[pagebackref,breaklinks,colorlinks]{hyperref}
\usepackage{url}            
\usepackage{booktabs}       
\usepackage{amsfonts}       
\usepackage{nicefrac}       
\usepackage{microtype}      
\usepackage{xcolor}         

\usepackage{graphicx}
\usepackage{amsmath}
\usepackage{multirow}
\usepackage{caption}
\usepackage{subcaption}
\usepackage{wrapfig}

\title{Uniform Masking: Enabling MAE Pre-training for Pyramid-based Vision Transformers with Locality}

\author{%
  Xiang Li$^{1}$, Wenhai Wang$^{2}$, Lingfeng Yang$^{1}$, Jian Yang$^{1}$\thanks{Corresponding author.}
  \\
  $^{1}$Nanjing University of Science and Technology, $^{2}$Shanghai AI Laboratory
  \\
  \\
  \texttt{\scriptsize \{xiang.li.implus, yanglfnju, csjyang\}@njust.edu.cn, wangwenhai362@163.com} \\
}

\begin{document}

\maketitle

\begin{abstract}
Masked AutoEncoder (MAE) has recently led the trends of visual self-supervision area by an elegant asymmetric encoder-decoder design, which significantly optimizes both the pre-training efficiency and fine-tuning accuracy. Notably, the success of the asymmetric structure relies on the \emph{``global''} property of Vanilla Vision Transformer (ViT), whose self-attention mechanism reasons over arbitrary subset of discrete image patches. However, it is still unclear how the advanced Pyramid-based ViTs (e.g., PVT, Swin) can be adopted in MAE pre-training as they commonly introduce operators within \emph{``local''} windows, making it difficult to handle the random sequence of partial vision tokens. In this paper, we propose Uniform Masking (UM) strategy, successfully enabling MAE pre-training for Pyramid-based ViTs with locality (termed ``UM-MAE'' for short). Specifically, UM includes a Uniform Sampling (US) that strictly samples $1$ random patch from each $2 \times 2$ grid, and a Secondary Masking (SM) which randomly masks a portion of (usually $25\%$) the already sampled regions as learnable tokens. US preserves equivalent elements across multiple non-overlapped local windows, resulting in the smooth support for popular Pyramid-based ViTs; whilst SM is designed for better transferable visual representations since US reduces the difficulty of pixel recovery pre-task that hinders the semantic learning. We demonstrate that UM-MAE significantly improves the pre-training efficiency (e.g., it speeds up by $\sim 2\times$ and reduces the GPU memory by at least $\sim 2\times$) of Pyramid-based ViTs, but maintains the competitive (or even better) fine-tuning performance across downstream tasks. For example using HTC++ detector, the pre-trained Swin-Large backbone \emph{self-supervised} under UM-MAE \emph{only in ImageNet-1K} can even outperform the one \emph{supervised in ImageNet-22K}.
The code and pre-trained models are available at https://github.com/implus/UM-MAE.



\end{abstract}


\begin{figure}[t]
	\vspace{0pt}
	\begin{center}
		\setlength{\fboxrule}{0pt}
		\fbox{\includegraphics[width=0.98\textwidth]{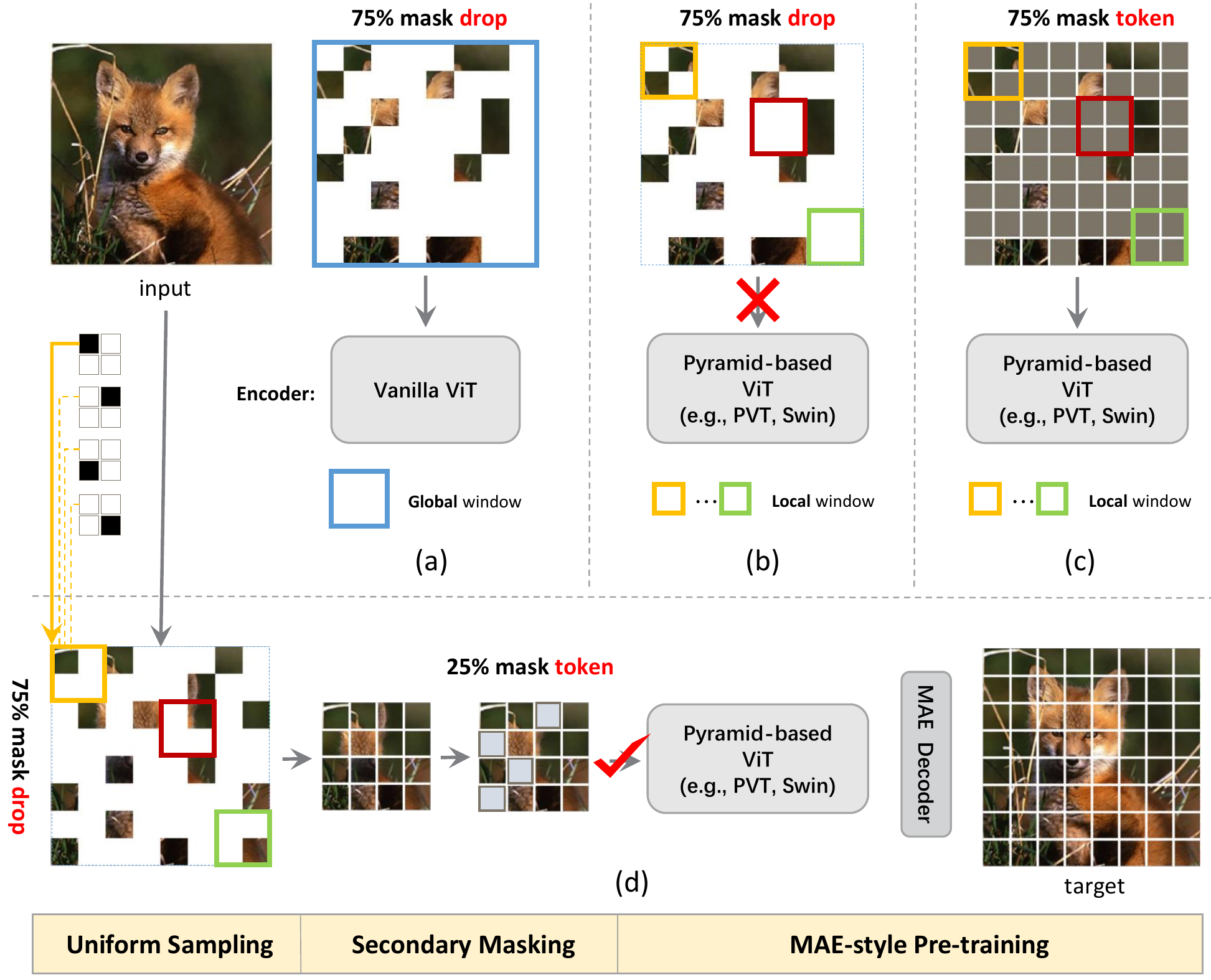}}
	\end{center}	
	\vspace{-9pt}
	\captionsetup{font={scriptsize}}
	\caption{\textbf{Illustration of various input strategy for supporting Vanilla ViT and Pyramid-based ViT.} (a) The ``global'' window of Vanilla ViT can receive arbitrary subset of image patches by skipping random $75\%$ of the total, whilst (b) skipping these $75\%$ patches is unacceptable for Pyramid-based ViT as patch elements are not equivalent across the ``local'' windows. (c) A straightforward solution is to adopt
    the mask token for the encoder (e.g.,  SimMIM~\cite{xie2021simmim}) at the cost of slower training. (d) Our Uniform Masking (UM) approach (including Uniform Sampling and Secondary Masking) enables the efficient MAE-style pre-training for Pyramid-based ViTs while keeping its competitive fine-tuning accuracy.
	}
	\label{fig_pipeline}
	\vspace{-10pt}
\end{figure}

\section{Introduction}
Self-Supervised Learning (SSL) uses auxiliary tasks to mine their own supervision from large-scale unlabeled data, and learns representations that are transferable to downstream tasks. SSL first shows great potential in Natural Language Processing (NLP) field by the ``masked autoencoding'' solutions of GPT~\cite{radford2018improving,radford2019language,brown2020language} and BERT~\cite{devlin2018bert}. These techniques, through learning to predict the removed portion of data based on the available context, have innovated the paradigm of NLP pipelines.

Inspired by the success of BERT, the vision community has recently raised great interest in imitating its formulation (i.e., masked autoencoding) for image understanding. A series of works~\cite{bao2021beit,dong2021peco,zhou2021ibot,chen2022context,he2021masked,xie2021simmim,wei2021masked,tong2022videomae} has been proposed in past months, where Masked AutoEncoder (MAE)~\cite{he2021masked} becomes one of the most representative methods which significantly optimizes both the pre-training efficiency and fine-tuning accuracy, successfully leading the new trend of SSL across vision tasks.

One of the most impactful designs in MAE is the asymmetric encoder-decoder architecture. Different from the decoder part where it receives entire sequence of patch tokens, the encoder part only takes the visible image patches (usually only 25\% of the total) as input. Interestingly, such design not only significantly reduces the pre-training complexity, but also obtains superior fine-tuning performance.

It is worth noting that the successful application of the asymmetric structure relies on the ``global'' property of Vanilla Vision Transformer (ViT)~\cite{dosovitskiy2020image}, whose self-attention mechanism can reason over arbitrary subset of discrete image patches (see Fig.~\ref{fig_pipeline}~(a)). Nevertheless, the ``global'' property of Vanilla ViT is a double-edged sword: When transferring Vanilla ViT to downstream vision tasks with considerably larger input resolution (e.g., $\sim 1000^2$ in object detection), its large memory requirement is challenging for modern GPUs due to the quadratic complexity of the global self-attention. Although ViTDet~\cite{li2022exploring,li2021benchmarking} attempts to partially restrict the local, windowed self-attention for certain blocks of Vanilla ViT during fine-tuning, the optimal architecture is unknown considering that the information flow can be arbitrarily different between pre-training and fine-tuning stage. Therefore, to address the above issue, the Pyramid-based ViTs~\cite{liu2021swin,liu2021swinv2,wang2021pyramid,wang2022pvtv2} can be the preferred options since they are more storage-friendly via naturally introducing the local window operators, and they have already demonstrated the great compatibility and advance in transferring to downstream vision tasks~\cite{khan2021transformers}. 

However, it is still unclear to effectively pre-train Pyramid-based ViTs using the efficient asymmetric structure likewise in MAE, as they are commonly equipped with operators in ``local'' windows. Specifically, the amount of visible elements in each local window is usually not equal, which prevents the effective parallel computation of window-based operators (see Fig.~\ref{fig_pipeline}~(b)). A compromise solution is to recall back the dropped patches as proposed in SimMIM~\cite{xie2021simmim}, where all the masked regions are modelled as learnable, shared mask tokens (see Fig.~\ref{fig_pipeline}~(c)). Consequently, it sacrifices a lot of efficiency due to the considerably large computation and storage complexity for the encoders.

To successfully enable MAE pre-training (i.e., adopt the efficient asymmetric structure) for Pyramid-based ViTs with locality, in this paper, we propose the Uniform Masking (UM) strategy that contains the Uniform Sampling (US) and Secondary Masking (SM) step. As illustrated in Fig.~\ref{fig_pipeline}~(d), US first strictly samples $1$ random patch from each $2 \times 2$ grid, resulting in $75\%$ drop of the image. Compared to the completely random $75\%$ drop in MAE, US has more risks to leak semantic clues as its sampled patches are distributed more uniformly than the Random Sampling (RS) of MAE, which reduces the difficulty of the pixel recovery pre-task (see the ``Pre-train Loss'' of  Table~\ref{table_vit_imagenet_ade20k_coco}) and hinders the representation learning (see Table~\ref{table_vit_imagenet_ade20k_coco}). Therefore, SM is further introduced to \emph{randomly} mask a (relatively small, e.g., $25\%$) portion
of the already sampled regions by US as shared, learnable tokens. Finally, the uniform-sampled patches, together with the secondary-masked tokens, are reorganized as a compact 2D input under a quarter of original image resolution, to sequentially feed through the Pyramid-based ViT encoder and MAE decoder for the reconstruction of the 
raw image pixels.

Our contributions are summarized as follows: 
(1) We propose Uniform Masking, which successfully enables MAE pre-training (i.e., UM-MAE) for popular Pyramid-based ViTs; 
(2) We empirically show that UM-MAE considerably speeds up pre-training efficiency by $\sim2\times$ and reduces the GPU memory consumption by at least $\sim2\times$ compared to the existing sota Masked Image Modelling (MIM) framework, whilst maintaining the competitive fine-tuning performance; 
(3) We reveal and discuss several notable different behaviors between Vanilla ViT and Pyramid-based ViTs under MIM.

\begin{figure}[t]
	\vspace{0pt}
	\begin{center}
		\setlength{\fboxrule}{0pt}
		\fbox{\includegraphics[width=0.98\textwidth]{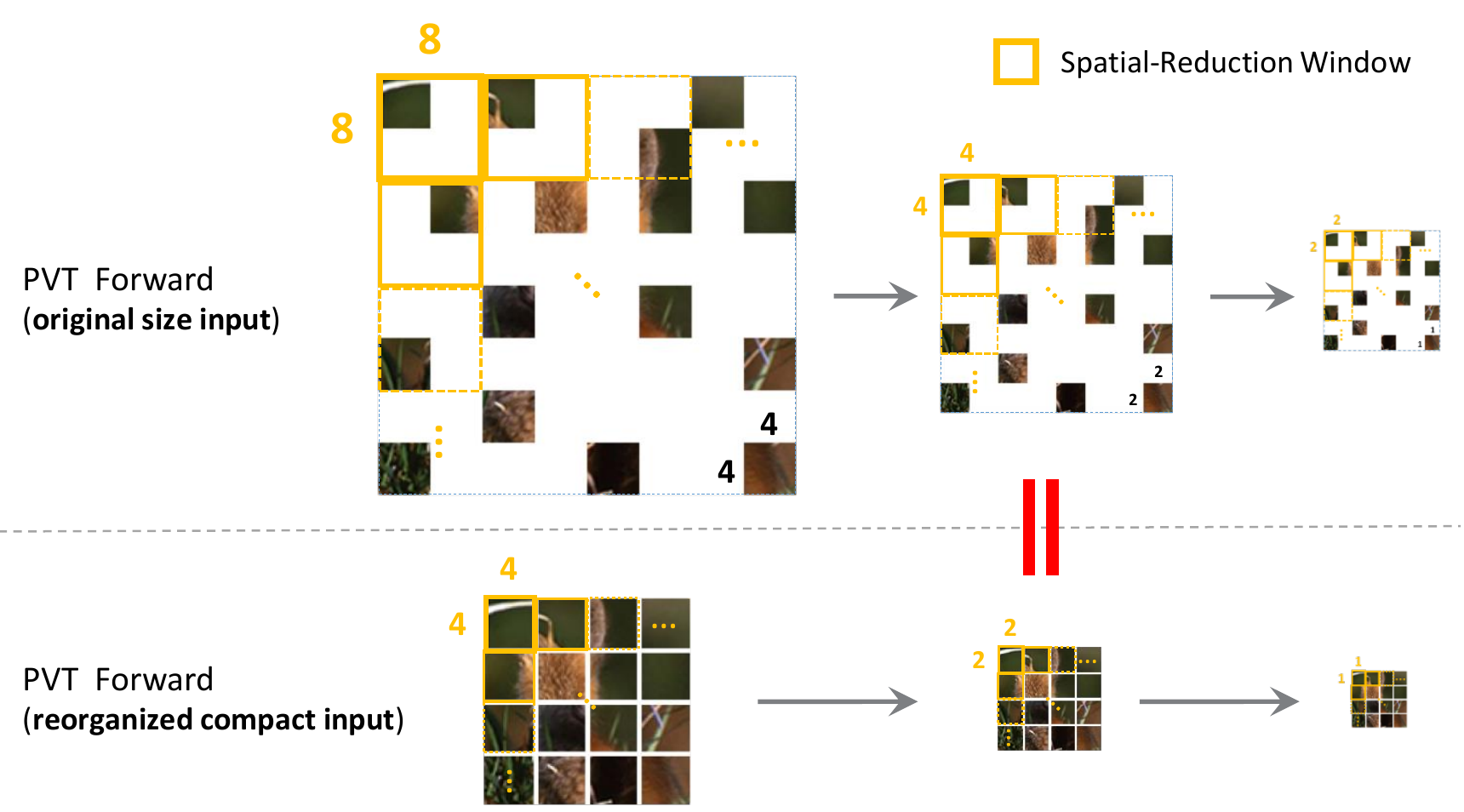}}
	\end{center}	
	\vspace{-9pt}
	\captionsetup{font={scriptsize}}
	\caption{\textbf{The equivalence between forwarding original size input with masked place holders and reorganized compact input for PVT~\cite{wang2021pyramid}.} We typical show three stages with pyramid spatial sizes. The spatial reduction window is performed over non-overlapped regions as in PVT.
	}
	\label{fig_pvt_equivalence}
	\vspace{-12pt}
\end{figure}

\section{Related Work}
\textbf{Masked Image Modeling (MIM).}
The past few months have witnessed a huge surge in MIM. BEiT~\cite{bao2021beit} first introduces the idea of BERT~\cite{devlin2018bert}, which dominates the field of Natural Language Processing (NLP), into computer vision. Due to the large structural difference between images and languages, they adopt the dVAE~\cite{ramesh2021zero} model to discretize image patches to obtain their class labels. The algorithm randomly masks the input image blocks, and expects the network to restore the categories of masked image patches  according to the context. To further improve the quality of the discretized labels, PeCo~\cite{dong2021peco} proposes to replace dVAE with VQ-VAE~\cite{van2017neural} model, and introduces perceptual loss in the reconstruction stage. iBOT~\cite{zhou2021ibot} and data2vec~\cite{baevski2022data2vec} attempt to use the mean teacher network to build an online discretization module, avoiding the additional and separate pre-steps of training the discretization model. CAE~\cite{chen2022context} proposes to separate representation learning from the image reconstruction task as much as possible, resulting in more robust feature representations.

Different from recovering discretized labels (i.e., high-dimensional semantic labels) of image patches, the Masked Autoencoder~\cite{he2021masked} (MAE) method is the first to demonstrate that high-quality image representations can also be obtained by directly recovering the original pixels of an image. MAE utilizes an asymmetric encoder-decoder design, that is, the encoder only receives visible image blocks as input in the feed-forward stage, which simultaneously improves the pre-training efficiency and achieves SOTA results on multiple image classification datasets. Concurrently, SimMIM~\cite{xie2021simmim} applies learnable mask tokens on the masked input patches and performs a similar self-supervised scheme to Swin Transformer~\cite{liu2021swin}, a Pyramid-based variant of ViT~\cite{dosovitskiy2020image}, achieving a competitive improvement. MaskFeat~\cite{wei2021masked} proves that the feature descriptor of the predicted image (e.g., HOG~\cite{dalal2005histograms}), instead of raw pixels, can further improve the fine-tuning performance.

\textbf{Pyramid-based Vison Transformer.} Ever since the successful application of Vanilla ViT~\cite{dosovitskiy2020image} in image classification, the vision community has put a lot of effort into researching its pyramid-based (i.e., hierarchical) variants~\cite{wang2021pyramid,wang2022pvtv2,yuan2021tokens,heo2021rethinking,wu2021cvt,liu2021swin,liu2021swinv2,srinivas2021bottleneck,chu2021twins,dong2021cswin,guo2021cmt,yang2021focal,fan2021multiscale,zhang2022nested,graham2021levit,zhang2021rest,gao2021container}, to inherit the advantages of the pyramid structure~\cite{lin2017feature} in the downstream vision tasks. There are two representative series among these variants: PVT~\cite{wang2021pyramid,wang2022pvtv2} and Swin~\cite{liu2021swin,liu2021swinv2}. Generally, PVT introduces non-overlapped Spatial-Reduction Window (SRW) to reduce the complexity of global self-attention mechanism, whilst Swin restrains self-attention operator inside non-overlapped, shifted local windows. Given their popularity, this paper focuses on these two representative architectures.

\begin{figure}[t]
	\vspace{0pt}
	\begin{center}
		\setlength{\fboxrule}{0pt}
		\fbox{\includegraphics[width=0.98\textwidth]{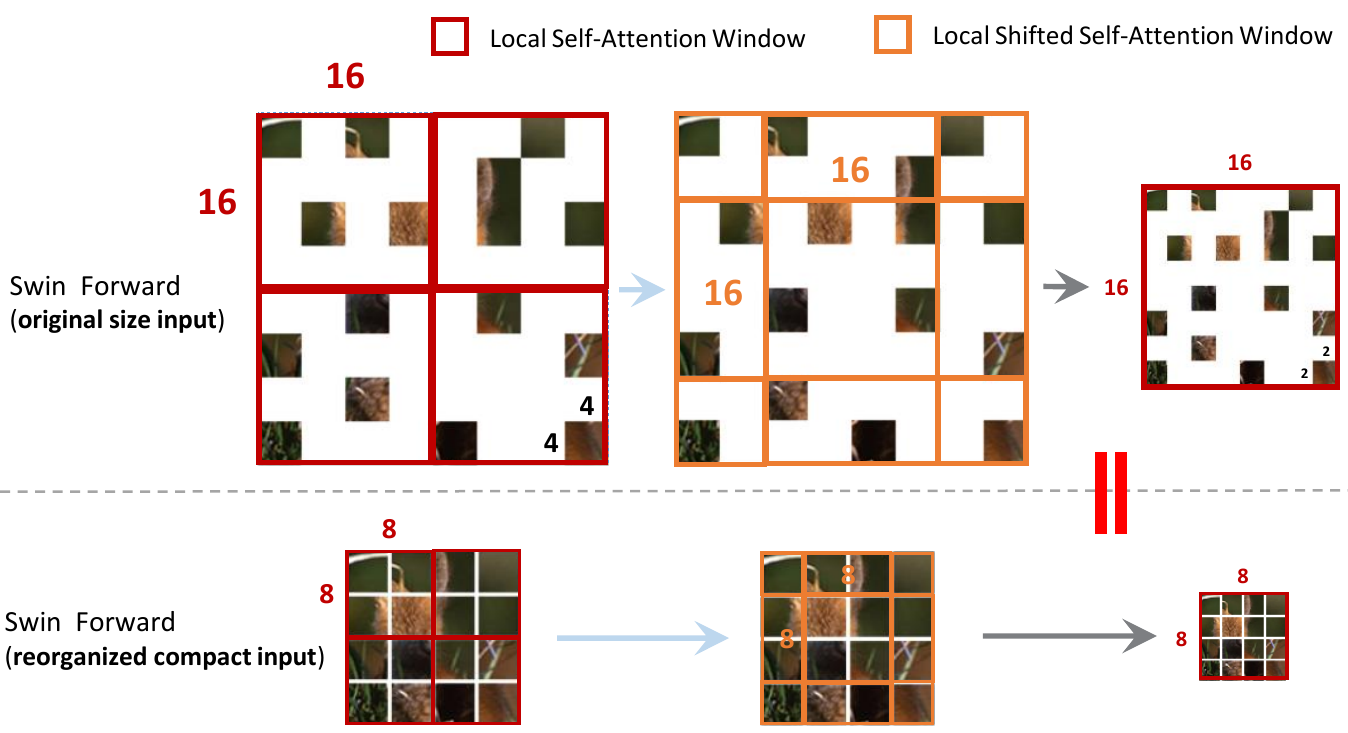}}
	\end{center}	
	\vspace{-9pt}
	\captionsetup{font={scriptsize}}
	\caption{\textbf{The equivalence between forwarding original size input with masked place holders and reorganized compact input for Swin~\cite{liu2021swin}.} We typically show the case of non-overlapped local self-attention window along with its shifted version in consecutive blocks of the same stage.
	}
	\label{fig_swin_equivalence}
	\vspace{-10pt}
\end{figure}

\section{Method}\label{sec:Method}
We propose Uniform Masking (UM) to enable the support of MAE-style pre-training for Pyramid-based ViTs.
UM is a simple and two-stage strategy that converts dense image tokens into sparse ones but spatially preserves their uniform distribution. It consists of two stages including Uniform Sampling (US) and Secondary Masking (SM), which together guarantee the efficiency and quality of self-supervised learning representations. We introduce them respectively in the following subsections.

\subsection{Uniform Sampling}
\label{sec_US}
As illustrated in Fig.~\ref{fig_pipeline}, UM firstly perform Uniform Sampling (US) that samples $25\%$\footnote{\scriptsize Here $25\%$ sample ratio empirically refers to the optimal mask ratio $75\%$ in MAE~\cite{he2021masked}.} visible image patches with a uniform constraint: $1$ of every $2 \times 2$ blocks is strictly sampled across the image space. Similarly following MAE~\cite{he2021masked}, the left $75\%$ masked patches are dropped and would not participate in the feed-forward process of the encoders.
US brings a desirable property to input data: the amount of elements is ensured equal across the shifted (and non-overlapped~\cite{wang2021pyramid,liu2021swin}) local windows. This property thus makes it possible for most popular Pyramid-based ViTs (e.g., PVT~\cite{wang2021pyramid} and Swin~\cite{liu2021swin}) to process the discrete subset of image patch sequence, as these sampled patches can be reorganized as a compact 2D image (see Fig.~\ref{fig_pipeline}~(d),~\ref{fig_pvt_equivalence},~\ref{fig_swin_equivalence}) 
thanks to the equivalence of the operated elements.
Taking the representative Pyramid-based ViTs: PVT~\cite{wang2021pyramid} and Swin~\cite{liu2021swin} as examples, we elaborate on the details of how US makes the uniform-distributed, sparse patches compatible for these architectures with locality. 

\textbf{Compatibility with PVT~\cite{wang2021pyramid}.} As shown in Fig.~\ref{fig_pvt_equivalence}, PVT introduces Spatial-Reduction Window (SRW) to reduce the complexity of the original self-attention block which builds relations over each feature pair. The information in SRW would be aggregated to present the keys and values with considerably reduced scales. In Fig.~\ref{fig_pvt_equivalence}, starting from the input image with uniformly sampled patches, we show the first three stages of PVT and mark its typical spatial-reduction hyper-parameters as $\{8, 4, 2\}$ sequentially in the upper pipeline. The masked patches are considered as blank place holders inside the original image/feature space. We demonstrate that for PVT, by simultaneously reorganizing the original visible input to its compact form and halving the edge sizes of spatial-reduction windows (i.e., $\{4, 2, 1\}$) accordingly, the effective elements are equivalent between corresponding local windows across the two pipelines. Therefore, US is compatible with PVT, and the input elements are reduced by $75\%$ in case of the encoder part when using the reorganized compact 2D image as input.

\textbf{Compatibility with Swin~\cite{liu2021swin}.} Swin has two major differences from PVT: (1) the Local Self-Attention Window (LSAW) size is usually fixed across stages; and (2) it has a Shifted version of the Local Self-Attention Window (SLSAW) in successive blocks, which introduces connections between neighboring non-overlapping windows. Based on these differences, we deduce that Swin has more restrictions on the selection of window size and input image scale during pre-training: a window (and input image) size of $16\cdot 2^n \times 16\cdot 2^n (n \in \mathbb{N})$ is necessary to ensure the equivalence when considering the shifted cases with a shift offset being $8\cdot 2^n \times 8\cdot 2^n$, as illustrated in Fig.~\ref{fig_swin_equivalence}. Under the above constraints, the effective elements are equivalent between corresponding (shifted) local windows across the two pipelines, similar to PVT.

\begin{figure}[t]
	\vspace{0pt}
	\begin{center}
		\setlength{\fboxrule}{0pt}
		\fbox{\includegraphics[width=0.98\textwidth]{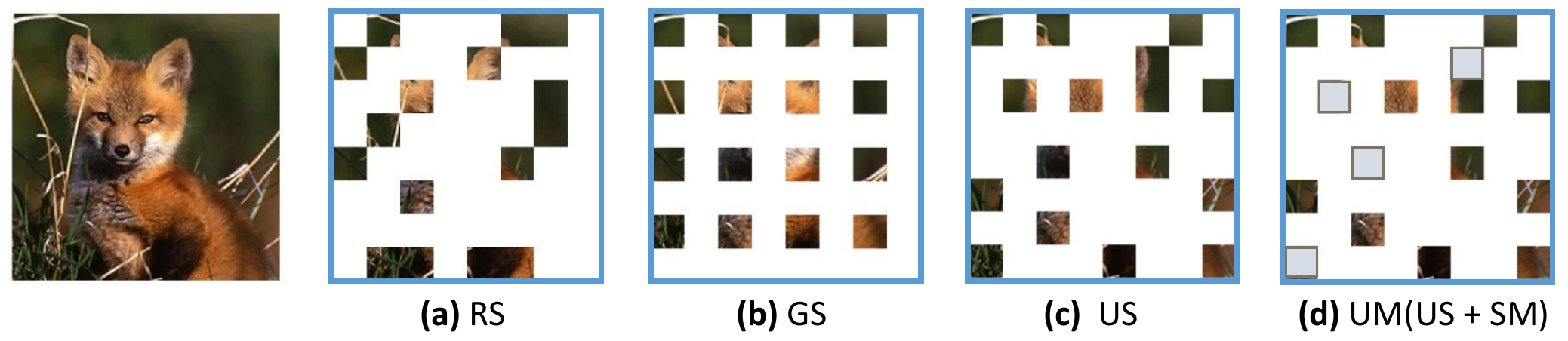}}
	\end{center}	
	\vspace{-10pt}
	\captionsetup{font={scriptsize}}
	\caption{\textbf{Different sampling strategy with a sample ratio of $25\%$.} (a) Random Sampling (RS) in MAE~\cite{he2021masked}; (b) Grid-wise Sampling (GS) in MAE; and the proposed (c) Uniform Sampling (US); (d) Uniform Masking (UM) that includes US and Secondary Masking (SM).
	}
	\label{fig_sampling_type}
	\vspace{-6pt}
\end{figure}

\subsection{Secondary Masking}
Compared to Random Sampling as adopted in MAE~\cite{he2021masked}, Uniform Sampling (US) samples the image patches uniformly distributed over the 2D space that makes it compatible for representative Pyramid-based ViTs. However, 
US can potentially make the self-supervisory task less challenging by providing shortcuts for pixel reconstruction purely through the neighbouring low-level image statistics.
It has already been evidenced in MAE, where a similar sampling strategy termed ``Grid-wise Sampling'' largely hinders the representation quality. It is also observed that in Table~\ref{table_vit_imagenet_ade20k_coco}, the uniform distribution by US indeed reduces the difficulty of the pre-training 
-- the pre-train loss decreases from 0.4256 (MAE Baseline) to 0.3858, but impedes the quality of learned representations for downstream tasks -- the fine-tuning accuracy is dropped by $\sim$1 absolute mIoU point in semantic segmentation.

To address the degradation issues brought by US, we further propose the Secondary Masking (SM) strategy that performs a secondary \emph{random} mask among the already sampled visible patches of US, as illustrated in Fig.~\ref{fig_sampling_type} from (c) to (d). Different from the US stage where the masked patches are entirely dropped, SM keeps the masked patches by using shared mask tokens for the compatibility of Pyramid-based ViTs with locality (see Sec.~\ref{sec_US}).
The simple operation thus increases the difficulty of semantic recovery pre-task, which focuses the network on learning high-quality representations over the incomplete visual context, without heavily relying on the neighbouring low-level pixels.

\subsection{UM-MAE Pipeline with Pyramid-based ViT}
\begin{figure}[t]
	\vspace{0pt}
	\begin{center}
		\setlength{\fboxrule}{0pt}
		\fbox{\includegraphics[width=0.98\textwidth]{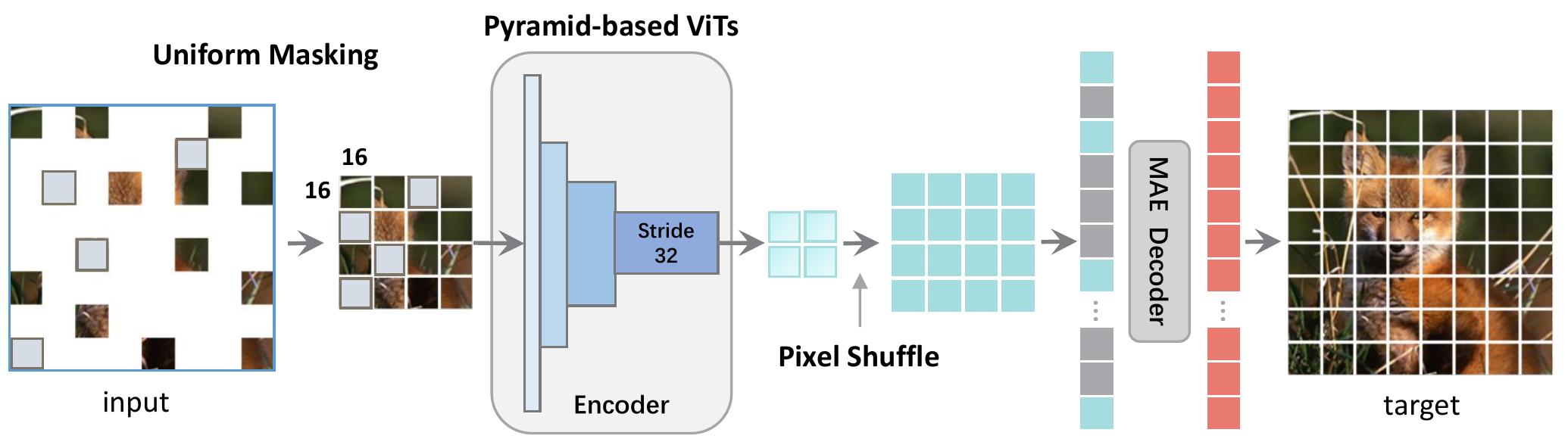}}
	\end{center}	
	\vspace{-7pt}
	\captionsetup{font={scriptsize}}
	\caption{\textbf{The Uniform Masking MAE pipeline (UM-MAE) for Pyramid-based ViTs.} The image patches are reorganized as 2D compact input after Uniform Masking as the input of Pyramid-based ViT Encoder (e.g., PVT, Swin). A Pixel Shuffle~\cite{shi2016real,paszke2019pytorch} layer is applied to recover the feature resolution due to the default large stride of the encoder backbone. The resulted representations are further fed into the MAE Decoder (lightweight Vanilla ViT) with mask tokens for the reconstruction of the original image.
	}
	\label{fig_detail_pipeline_cropped}
	\vspace{-4pt}
\end{figure}

Fig.~\ref{fig_detail_pipeline_cropped} demonstrates the detailed asymmetric design (i.e., the MAE-style pipeline) of our approach when applied to typical Pyramid-based ViTs (e.g., PVT and Swin). For easier reference, we term our method as UM-MAE and describe its major components as follows.

\textbf{Encoder.} We follow MAE~\cite{he2021masked} where the operation units are $16\times 16$ image patches. The proposed Uniform Masking is performed to obtain the compact, reorganized 2D input (consisting of both visible patches and mask tokens). It is with reduced scale (i.e., $25\%$ of the full set) as the input for the encoder. Each mask token~\cite{devlin2018bert} is a shared, learned embedding vector that indicates the presence of a miss. Whether to append the positional embeddings here is decided by the specific architectures used in the encoder (e.g., ``yes'' for PVT and ``no'' for Swin by default). 
However, hierarchical ViTs usually have four stages with a total stride being 32, which makes the $16\times 16$ unit become a sub-pixel finally. Therefore, the sub-pixel convolution~\cite{shi2016real} is adopted (i.e., PixelShuffle operator in Pytorch~\cite{paszke2019pytorch}) to recover its resolution before feeding the learned representations to the decoder.

\textbf{Decoder.} The decoder part is similar to that of MAE. The input to the decoder is the full set of tokens including (i) encoded patches from UM, and (ii) mask tokens. Positional embeddings are added to all tokens and the architecture of decoder refers to the lightweight Vanilla ViT as adopted in MAE.

\textbf{Reconstruction Target.} We reconstruct the input by predicting the pixel values (the normalized version in MAE) for each \emph{dropped} patch during Uniform Masking. A linear projection is applied on the last layer of the decoder whose number of elements equals the number of pixel values in a patch. The loss function is Mean Squared Error (MSE) and it is computed only on the \emph{dropped} patches. 

\section{Experiment}
In this paper, as suggested in MAE~\cite{he2021masked}, we focus on the fine-tuning accuracy (instead of linear probing) over a series of downstream tasks, including image classification using ImageNet-1K (IN1K)~\cite{deng2009imagenet}, semantic segmentation using ADE20K~\cite{zhou2017scene} and object detection using COCO~\cite{lin2014microsoft}. The pre-training is conducted on IN1K training set with $256^2$ resolution by default. Unless otherwise stated, we perform supervised fine-tuning (or learning from scratch) over image classification for 100 epochs with $256^2$ resolution, semantic segmentation for 160K iterations with $512^2$ resolution, and object detection for 25 epochs with $1024^2$ resolution. We adopt UperNet~\cite{xiao2018unified} as the segmentor and GFL~\cite{li2020generalized} as the object detector. The depth of decoder in UM-MAE is set to 2 for faster ablation iterations. 
To show the efficiency and effectiveness of our method, we mainly compare against the existing sota self-supervised approach, i.e., SimMIM~\cite{xie2021simmim} that has already been adopted successfully in pre-training Pyramid-based ViT like Swin~\cite{liu2021swin,liu2021swinv2}. Two representative models from Pyramid-based ViTs, PVT-Small (PVT-S) and Swin-Tiny (Swin-T) are adopted in most ablation studies for experimental efficiency. Our 
codebase is built upon the official repositories of MAE, SimMIM, MMDetection and MMSegmentation\footnote{\scriptsize https://github.com/facebookresearch/mae, https://github.com/microsoft/SimMIM, https://github.com/open-mmlab}. More details refer to the Supplementary Material.

\subsection{Pilot Experiments with Vanilla ViT}
Instead of directly applying UM to Pyramid-based ViTs, we first conduct pilot experiments using UM strategy in pre-training Vanilla ViT to validate its effectiveness and capture a brief understanding of its mechanism. It is believed that ``UM works for Vanilla ViT'' is a necessary (but not sufficient) condition for ``UM works for Pyramid-based ViTs''. 

\textbf{Pilot Settings.} ViT-Base (ViT-B/16)~\cite{dosovitskiy2020image} is adopted as the backbone in our pilot experiments. Following the official codes of MAE, we first do self-supervised pre-training (under 200 epochs) on the ImageNet-1K (IN1K)~\cite{deng2009imagenet} training set using different sampling strategies as illustrated in Fig.~\ref{fig_sampling_type}. For fair comparisons, the number of input image patches and self-supervised signals are kept \emph{exactly the same} across the sampling strategies during pre-training. Then we perform supervised fine-tuning over image classification (100 epochs, $224^2$ resolution), semantic segmentation (160K iterations, $512^2$ resolution) and object detection (25 epochs, $896^2$ resolution). We adopt UperNet~\cite{xiao2018unified} as the segmentor and GFL~\cite{li2020generalized} as the object detector.
The other pre-training/fine-tuning hyper-parameters follow the descriptions from MAE~\cite{he2021masked}. More details can be found in Supplementary Material. For the mask ratio of Secondary Masking in UM, we ablate their values in $\{15\%, 25\%, 35\%\}$.

\textbf{Results.} In Table~\ref{table_vit_imagenet_ade20k_coco}, it is observed that the proposed UM with $25\%$ SM Ratio is as competitive as the RS (MAE Baseline), whilst other variants lead to performance degradation during fine-tuning. The results of (a), (b), (c) show the more uniform the distribution, the simpler the pre-training task and the worse the transfer ability. Our proposed UM (with $25\%$ SM Ratio) achieves the best trade-off.

\begin{table}
	\vspace{0pt}
    \renewcommand\arraystretch{1.2}
    \setlength{\tabcolsep}{4.pt}
    \footnotesize
    \centering
    \resizebox{0.96\textwidth}{!}
    {
        \begin{tabular}{l|c|c|c|c|cc|ccc}
        \hline
        \multirow{2}{*}{Sampling Strategy ($25\%$)} & Pyramid & SM & Pre-train & {ImageNet-1K} & \multicolumn{2}{c|}{ADE20K} & \multicolumn{3}{c}{COCO}  \\
        \cline{5-10}
        & Support & Ratio & Loss & Top-1 Acc  & mIoU & aAcc & AP & AP$_{50}$ & AP$_{75}$ \\
        \hline
        (a) RS (MAE~\cite{he2021masked} Baseline)& $\times$ & -- & 0.4256
 &  \textbf{82.88} 
 &  \textbf{42.54} & \textbf{80.85} & \textbf{46.0} & \textbf{64.7} & \textbf{49.8}  \\
        
        (b) GS  & \checkmark  & -- & 0.3682 & 82.48 &  38.79 & 79.16 &  44.4 & 63.2 & 48.6 \\
        
        (c) US (Ours) & \checkmark & -- & 0.3858 & 82.74 & 41.55
 & 80.48 & 45.5 & 64.2 & 49.6\\
        \hline
        \multirow{3}{*}{(d) UM (Ours)}& \checkmark & $15\%$ & 0.4171
 & 82.75 & 41.68 & 80.54 & 45.8 & \textbf{64.6} & 49.8 \\
        
         & \checkmark & $25\%$ & 0.4395
 & \textbf{82.88} & \textbf{42.59} & \textbf{80.80} & \textbf{45.9} & 64.5 & \textbf{50.2}  \\
        
         & \checkmark & $35\%$ & 0.4645 & 82.68 & 42.02 & 80.72 & \textbf{45.9} & \textbf{64.6} & 50.1\\
        \hline 
        
        \end{tabular}
    }
	\vspace{2pt}
	\captionsetup{font={scriptsize}}
    \caption{\textbf{Comparisons among different sampling strategies using Vanilla ViT-Base (ViT-B/16) backbone under 200-epoch pre-training.} ``Pyramid Support'' denotes whether the strategy supports the Pyramid-based ViTs. ``SM Ratio'' denotes the Secondary Masking Ratio. ``Pre-train Loss'' refers to the converged pre-training loss value which can represent the difficulty of the self-supervisory task.
    }
    \label{table_vit_imagenet_ade20k_coco}
	\vspace{-14pt}
\end{table}

\subsection{Experiments with Pyramid-based ViT}
\begin{wraptable}{r}{0.4\textwidth}
	\vspace{-10pt}
    \input{./wrap/table_sm_ratio.tex}
	\vspace{12pt}
\end{wraptable}
\textbf{Ablation Study on Secondary Masking Ratio.} According to the optimal mask ratio (i.e., $25\%$) validated in the pilot experiment using Vanilla ViT, we continue to perform ablation study on Secondary Masking Ratio around $25\%$ (i.e., $\{20\%, 25\%, 30\%\}$) for Pyramid-based ViTs (here we take Swin-T as an example), which are the focus of this paper. The models fine-tuned on ImageNet are also used for fine-tuning on COCO object detection and ADE20K semantic segmentation tasks, following the intermediate fine-tuning scheme as proposed in BEiT~\cite{bao2021beit} and adopted in SwinV2~\cite{liu2021swinv2}. The pre-training schedule is set as 200 epoch for efficiency. As shown in Table~\ref{table_table_sm_ratio}, comprehensively considering the performance of multiple downstream tasks, a good choice of SM Ratio for Pyramid-based ViT is $25\%$, as same as that of Vanilla ViT. The ratio is thus fixed as $25\%$ by default in the following experiments.

\textbf{Ablation Study on Pre-training Schedules.} We ablate the pre-training schedules of both UM-MAE (with $25\%$ SM Ratio) and SimMIM (with the optimal $25\%$ RS as in MAE) from 200 to 800 to show its influence on the IN1K fine-tuning accuracy using PVT-S. We fix the fine-tuning epoch as 100. The supervised baseline of learning from ``Scratch'' for 100 epochs is also provided for a better comparison. Fig.~\ref{fig_pretrain_epoch} shows that more pre-training epochs lead to better fine-tuning accuracy.

\textbf{Ablation Study on Fine-tuning Schedules.} We empirically discover that the self-supervised Pyramid-based ViTs using MIM can have consistent benefits from more fine-tuning epochs. Based on PVT-S which is first pre-trained using SimMIM and UM-MAE on ImageNet-1K for 200 epochs, we fine-tune the model again on ImageNet-1K for 100, 200 and 300 epochs, respectively. The direct supervised baseline (i.e., learning from ``Scratch'') is also provided. Fig.~\ref{fig_finetune_epoch} demonstrates that UM-MAE performs on par with SimMIM, and they both outperform the supervised baseline by a non-trivial margin.

\begin{figure}[t]
    \vspace{0pt}
    \centering
    \input{wrap/figure_sm_ratio_pretrain_epoch.tex}
    \vspace{0pt}
\end{figure}

\textbf{Efficiency.} Compared to the SimMIM framework for self-supervising Pyramid-based ViTs, the core advantage of the proposed UM-MAE is the memory and runtime efficiency. In Table~\ref{table_efficiency}, we show their clear comparisons based on PVT-S and Swin-T in case of pre-training over 200 epochs. The fine-tuning and supervised-from-scratch performances are as well demonstrated. The statistics of pre-training time and memory usage are evaluated on 8 GeForce RTX 3090 GPUs. It is observed that the proposed UM-MAE significantly saves almost \emph{half} of the pre-training hours and around { $\frac{1}{2} $ to $ \frac{2}{3}$ GPU memory budgets} against SimMIM, where their performances under multiple downstream tasks are comparable (and sometimes better, e.g., 45.96 vs. 45.35 mIoU on ADE20K).

\begin{table}
	\vspace{0pt}
    \renewcommand\arraystretch{1.2}
    \footnotesize
    \centering
    \resizebox{0.92\textwidth}{!}
    {
        \begin{tabular}{l|l||c|c||l|l|l}
        \hline
        \multirow{2}{*}{Architecture}& \multirow{2}{*}{Method} & \multicolumn{2}{c||}{Pre-train { (200 epoch)} } & \multicolumn{3}{c}{Fine-tune (/Scratch) Performance}
        \\\cline{3-7}
         & &  Time & Memory & ImageNet-1K &  ADE20K & COCO \\
        \hline
        \multirow{3}{*}{PVT-S~\cite{wang2021pyramid}} & \multicolumn{3}{l||}{Supervised from Scratch (Baseline)}  & 77.84 & 40.38 & 42.3 \\ \cline{2-7}
        & SimMIM~\cite{xie2021simmim} &  38.0 h & 20.6 GB & 79.28 {\scriptsize (+1.44)} & \textbf{43.04 {\scriptsize (+2.66)}} & 44.8 {\scriptsize (+2.5)}  \\
        & UM-MAE \textbf{(ours)} & \textbf{21.3 h}  & \textbf{11.6 GB} & \textbf{79.31 {\scriptsize (+1.47)}} & 43.01 {\scriptsize (+2.63)} & \textbf{45.1 {\scriptsize (+2.8)}} \\ \hline
        
        \multirow{3}{*}{Swin-T~\cite{liu2021swin}} & \multicolumn{3}{l||}{Supervised from Scratch (Baseline)}  & 81.82 & 44.51 & 47.2 \\ \cline{2-7}
        & SimMIM~\cite{xie2021simmim} & 49.3 h  & \ \ 37.4 GB* & \textbf{82.20 {\scriptsize (+0.38)}}  & 45.35 {\scriptsize (+0.84)} & 47.6 {\scriptsize (+0.4)}\\
        & UM-MAE \textbf{(ours)} & \textbf{25.0 h}  & \textbf{13.4 GB} & 82.04 {\scriptsize (+0.22)} & \textbf{45.96 {\scriptsize (+1.45)}} & \textbf{47.7 {\scriptsize (+0.5)}} \\ \hline
        \end{tabular}
    }
	\vspace{2pt}
	\captionsetup{font={scriptsize}}
    \caption{\textbf{Efficiency comparisons with representative fine-tuning performance.} The real time (hours) and memory consumption (GB) are evaluated on 8 GeForce RTX 3090 GPUs with 128 images per GPU ($*$ denotes the estimation due to memory limitation 24 GB). To train Swin-T under SimMIM which exceeds the GPU memory, we halve the batch size per GPU and double the accumulation step for maintaining the same effective global batch size. The models fine-tuned on ImageNet are also used for fine-tuning on COCO object detection and ADE20K semantic segmentation (i.e., the intermediate fine-tuning scheme~\cite{bao2021beit,liu2021swinv2}). Compared to SimMIM, UM-MAE speeds up by $\sim 2\times$ and reduces the memory by at least $\sim 2\times$, whilst performing competitively.
    }
    \label{table_efficiency}
	\vspace{-20pt}
\end{table}

\begin{wraptable}{r}{0.42\textwidth}
    \input{./wrap/table_large_model.tex}
	\vspace{12pt}
\end{wraptable}

\textbf{Performance on Large Models.} The above ablation study mainly focuses on relatively small models for experimental efficiency. We are interested whether the proposed approach can scale to large architectures, e.g., Swin-Large (Swin-L)~\cite{liu2021swin}. Following the settings of SimMIM~\cite{xie2021simmim}, we apply UM-MAE pre-training for 800 epochs and perform fine-tuning for 100 epochs on IN1K. 
Table~\ref{table_large_model} reports the Top-1 accuracy, showing that our UM-MAE maintains the competitive performance on large-scale models. The model is further fine-tuned on COCO under the same HTC++~\cite{chen2019hybrid,liu2021swin} framework and evaluated on COCO validation set. Table~\ref{table_large_model_coco} demonstrates that the pre-trained backbone self-supervised under UM-MAE \emph{only in IN1K} can even outperform the one supervised in IN22K, 
using \emph{only half} the training epochs of Baseline. 

\begin{figure}[t]
	\vspace{0pt}
	\begin{center}
		\setlength{\fboxrule}{0pt}
		\fbox{\includegraphics[width=0.98\textwidth]{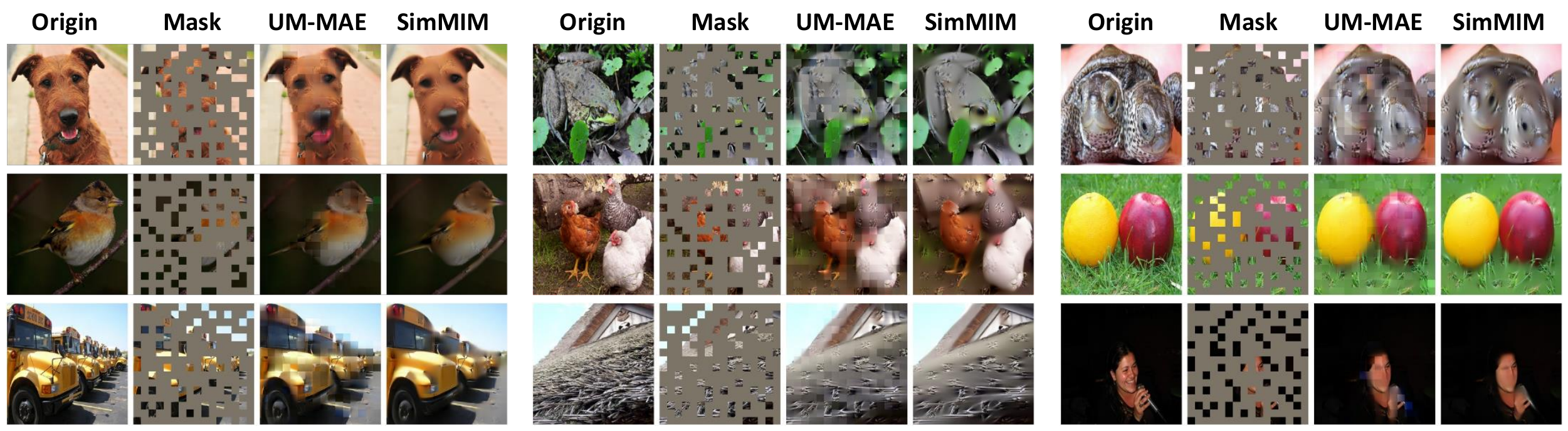}}
	\end{center}	
	\vspace{-15pt}
	\captionsetup{font={scriptsize}}
	\caption{\textbf{Uncurated reconstruction visualizations under the same $75\%$ mask pattern.} The models are both pre-trained for 800 epochs.
	}
	\label{fig_visual_reconstruction_cropped}
	\vspace{-7pt}
\end{figure}

\begin{wraptable}{l}{0.49\textwidth}
    \input{./wrap/table_large_model_coco.tex}
	\vspace{12pt}
\end{wraptable}
\textbf{Visualization of Reconstructions.} Based on Swin-T, we visualize the image reconstruction results for both UM-MAE and SimMIM, using a shared mask drawn from Uniform Sampling. Note that for UM-MAE we further perform Secondary Masking at a ratio of $25\%$ as input mask embeddings for final reconstruction (i.e., UM-MAE has $18.75\%$ valid input image patches in fact). In Fig.~\ref{fig_visual_reconstruction_cropped}, we observe that they all roughly recover the semantic details close to the original images, whilst the results of SimMIM may be over-smooth sometimes, e.g., the \emph{long} right leg of the frog. 

\subsection{Discussion: Different Behaviors between Vanilla ViT and Pyramid-based ViT under MIM} 
For downstream tasks with dense predictions, we empirically discover several notable differences between Vanilla ViT and Pyramid-based ViT under the framework of MIM. Take the semantic segmentation task on ADE20K as an example, we mainly demonstrate the following two aspects:

\textbf{(1) The performance of direct fine-tuning self-supervised Pyramid-based ViTs under MIM lags far behind the intermediate fine-tuning scheme as proposed in BEiT~\cite{bao2021beit}.} From Table~\ref{table_discussion_intermediate}, we observe that the self-supervised Pyramid-based ViTs under MIM framework are not competitive in directly fine-tuning downstream tasks with dense predictions. However, an intermediate fine-tuning process on IN1K can significantly improve its performance, which even outperforms the supervised baseline by a considerable margin (44.51 $\to$ 45.96). Following BEiT~\cite{bao2021beit}, the intermediate fine-tuning further trains the model on an intermediate dataset after MIM pre-training, and then fine-tunes the model on the target downstream tasks. The similar scheme is also supported in SwinV2~\cite{liu2021swinv2}, where all the demonstrated top results of Swin Transformers on downstream dense prediction tasks contain an intermediate fine-tuning step, instead of directly fine-tuning the model self-supervised by SimMIM~\cite{xie2021simmim} on the target tasks. This phenomenon is quite different from the case of Vanilla ViT, where the self-supervised Vanilla ViT after MIM pre-training can directly beats the supervised counterparts (e.g. on ADE20K,  45.3 $\to$ 45.6 in BEiT~\cite{bao2021beit} and 47.4 $\to$ 48.1 in MAE~\cite{he2021masked}).
Considering the lack of intrinsic inductive bias in modeling local visual structures, 
we deduce that the {Vanilla ViT} can benefit \emph{much more} from the pre-training MIM stage as it usually suffers from many optimization problems (e.g., attention collapse~\cite{zhou2021deepvit}). MIM tends to provide a fantastic initialization and probably brings effective inductive bias (see the next paragraph) for Vanilla ViT through meaningful semantic reconstruction over the visual context. Therefore, it has a great chance to avoid the bad local minimums and fully unleash the potential of Vanilla ViT during supervised fine-tuning on downstream tasks with dense predictions.

\textbf{(2) Layer-wise learning rate decay~\cite{clark2020electra,bao2021beit} is crucial for Vanilla ViT to obtain the optimal performance, but is harmful for Pyramid-based ViTs.} As demonstrated in Table~\ref{table_discussion_lr_decay}, when fine-tuning models on ADE20K, Vanilla ViT relies heavily on the layer-wise learning rate decay strategy, whilst Pyramid-based ViTs have the opposite effect. We suspect that the pre-training MIM step probably brings effective inductive bias for Vanilla ViT especially under its early layers, thereby a relatively small learning rate at early stages would not dramatically change the pre-learned implicit local structures which preserve the generalization ability. In contrast, Pyramid-based ViTs have already introduced the inductive bias via local window operators, thus there is probably no need for them to apply the layer-wise learning rate decay strategy in this case.


\begin{table}[t]
    \vspace{0pt}
    \centering
    \input{wrap/table_discussion.tex}
    \vspace{-4pt}
\end{table}

\section{Conclusion}
In this paper, we propose UM-MAE with a novel Uniform Masking (UM) strategy which contains a Uniform Sampling (US) and a Secondary Masking (SM) step, successfully enabling MAE pre-training for popular Pyramid-based Vision Transformers with locality. Compared to the existing alternative SimMIM, UM-MAE significantly improves the pre-training efficiency in both memory and runtime of Pyramid-based ViTs but maintains the competitive fine-tuning performance. We also discuss several empirical findings about the different behaviors between Vanilla ViT and Pyramid-based ViT under the framework of MIM. We hope our approach and discovery can faithfully inspire the vision community and push forward the topic of MIM.







{
\small
\bibliographystyle{plain}
\bibliography{egbib}
}

\appendix

\begin{table}
	\vspace{0pt}
    \renewcommand\arraystretch{1.2}
    \footnotesize
    \centering
    \resizebox{0.96\textwidth}{!}
    {
        \begin{tabular}{l|l|l}
        \hline
config & Pre-train Value & Fine-tune Value \\\hline
optimizer & AdamW  & AdamW \\
base learning rate & 1.5e-4 & 5e-4\\
weight decay & 0.05 & 0.05 \\
optimizer momentum & $\beta_1, \beta_2{=}0.9, 0.95$  & $\beta_1, \beta_2{=}0.9, 0.999$ \\
layer-wise lr decay & 1.0 & 0.7 (Swin-L), 0.8 (ViT-B), 0.85 (Swin-T, PVT-S) \\
global batch size (over 8 GPUs) & 4096 & 1024 \\
batch size per GPU & 128 & 64\\
accumulated iteration & 4 & 2 \\
learning rate schedule & cosine decay & cosine decay\\
warmup epochs  & 10 & 5 \\
augmentation & RandomResizedCrop &  RandAug (9, 0.5)\\
label smoothing & -- & 0.1 \\
mixup  & -- & 0.8 \\
cutmix  & -- & 1.0 \\
drop path & -- & 0.1 (ViT-B, Swin-T, PVT-S), 0.2 (Swin-L) \\
\hline
        \end{tabular}
    }
	\vspace{2pt}
	\captionsetup{font={scriptsize}}
    \caption{Pre-training and fine-tuning settings on ImageNet-1K dataset.
    }
    \label{table_sm_pretrain_finetune_IN1K}
	\vspace{-16pt}
\end{table}

\section{Broader Impacts}
The proposed method may generate inexistent images and introduce biases from the learned training dataset, which would contain possible negative societal impacts. These issues warrant further research and consideration when using the pre-trained models.

\section{Implementation Details}
\subsection{Architectures}
\textbf{Vanilla ViT Architecture.} We follow the standard Vanilla ViT architecture adopted in MAE~\cite{he2021masked}. The encoder and decoder share the similar structures which consist of a stack of Transformer blocks. The positional embeddings use the sine-cosine version as well. To fine-tune image classification tasks with Vanilla ViT, we extract globally averaged feature vector from the encoder output.

\textbf{PVT Architecture.} The PVT architecture generally follows the first version of PVT~\cite{wang2021pyramid}. The positional embeddings use the sine-cosine version as well, and they are adopted only in the input stage. In order to flexibly adjust the operation size of SRW (e.g., from window size 4 to 8) when transferring to downstream tasks, we adopt the spatial pooling operator as proposed in \cite{wang2022pvtv2} for the implementation of SRW. The architectures and pre-training/fine-tuning resolutions are kept the same under different MIM methods in our ablation experiments for fair comparisons.

\textbf{Swin Architecture.} The Swin architecture generally follows the first version of Swin~\cite{liu2021swin}. To flexibly transfer the relative position biases to downstream tasks with arbitrary input scales, we refer to the continuous approach of \cite{liu2021swinv2} where a small meta network is adopted to produce the biases based on the log-spaced relative coordinates. The architectures and pre-training/fine-tuning resolutions are kept the same under different MIM methods in our ablation experiments for fair comparisons.

\subsection{Pre-training and Fine-tuning Settings on ImageNet-1K}
ImageNet-1K~\cite{deng2009imagenet} contains 1.3M images of 1K categories for training and 50K images for validation. The larger dataset, ImageNet-22K (IN22K), which contains 14.2M
images and 22K classes, is not used at all in our method. 
In this paper, the pre-training is conducted on ImageNet-1K training set. The default setting is in Table~\ref{table_sm_pretrain_finetune_IN1K} in case of pre-training for 200 epochs, where most of the configuration values refer to that of MAE~\cite{he2021masked}. The warmup epoch is linearly scaled when the pre-training epoch increases to 400 or 800. To train the models which exceeds the GPU memory, we halve the batch size per GPU (e.g., 128 $\to$ 64) and double the accumulation step (e.g., 4 $\to$ 8) for maintaining the same effective global batch size. We follow MAE and use the linear \textit{lr} scaling rule: \textit{lr} = \textit{base\_lr}$\times$globalbatchsize / 256.

\begin{figure}[t]
	\vspace{0pt}
	\begin{center}
		\setlength{\fboxrule}{0pt}
		\fbox{\includegraphics[width=0.98\textwidth]{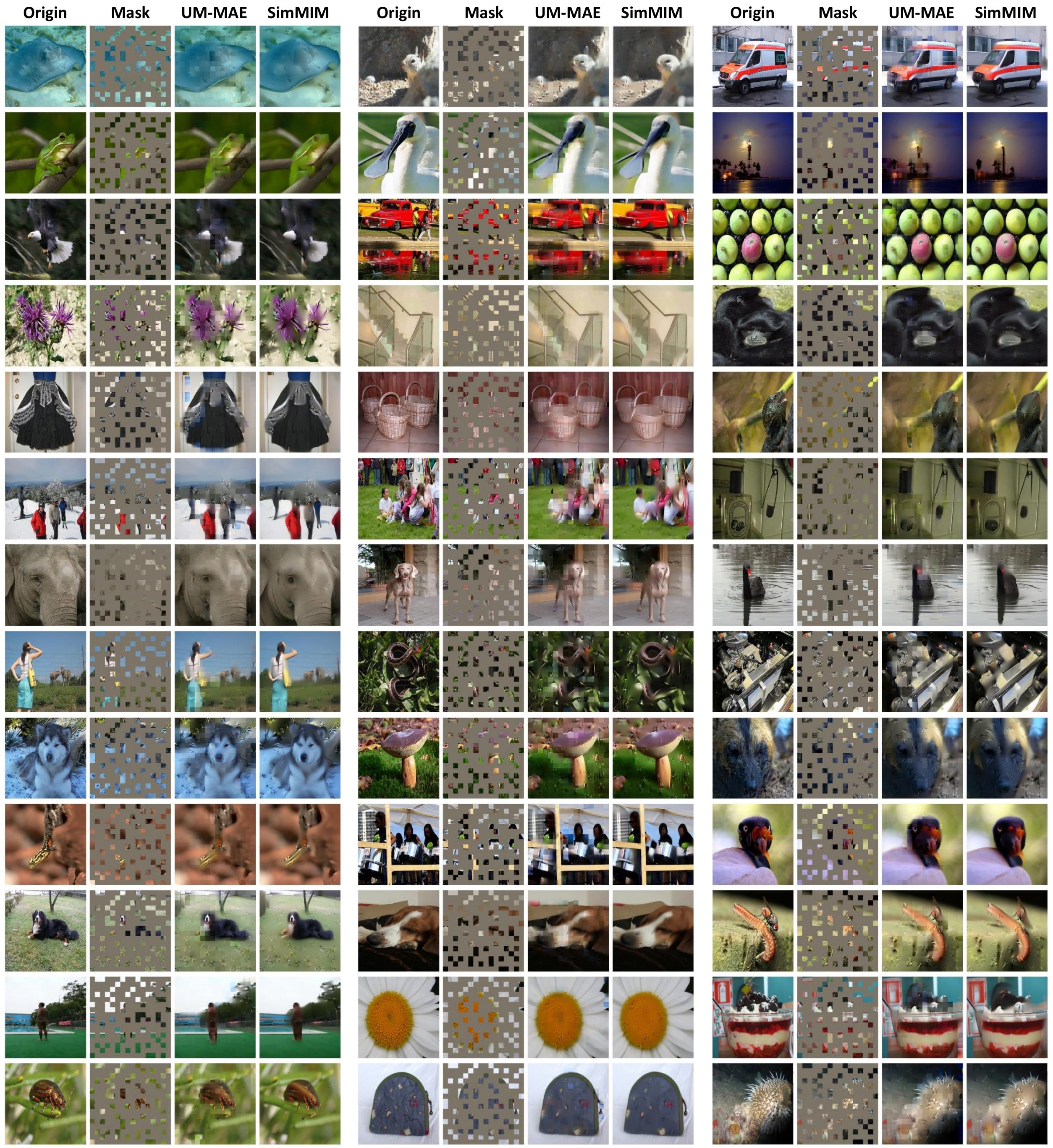}}
	\end{center}	
	\vspace{-14pt}
	\captionsetup{font={scriptsize}}
	\caption{\textbf{Uncurated random samples} on ImageNet \textit{validation} images. The masking ratio is $75\%$. The Swin-T models under UM-MAE and SimMIM are both pre-trained for 800 epochs.
	}
	\label{fig_sm_more_visual_cropped}
	\vspace{-4pt}
\end{figure}

\subsection{Object Detection on COCO}
The COCO benchmark~\cite{lin2014microsoft} consists of  trainval35k (115K images) for training and minival (5K images) for validation. We adopt the one-stage GFL~\cite{li2020generalized} detector on COCO dataset. Specifically, we replace the backbone of GFL with the pre-trained Vanilla ViT, PVT and Swin, respectively. The hyper-parameters mainly follow the default settings of GFL for PVT in mmdetection~\cite{chen2019mmdetection}. We report the performance using AdamW optimizer under a 25 epochs cosine schedule with a base learning rate of 1e-4, a weight decay of 0.05, and augmentation by large-scale jitter with a scale range of [0.1, 2.0]. The batch size is 16, distributed across 8 GPUs (i.e., 2 images per GPU).

\textbf{For Vanilla ViT (ViT-B in this paper),} we use the local-global adaptation during fine-tuning as proposed in \cite{li2021benchmarking}, where the network blocks are divided into 4 subsets, each consisting of a last global-window block and several local-window blocks otherwise. The intermediate feature maps at the tail of each subset are applied with convolutions to upsample or downsample to different scales for matching the input requirement of FPN~\cite{lin2017feature}. The default training resolution is $1024^2$ as suggested in \cite{li2021benchmarking,li2022exploring}. However, even under the memory-friendly implementation of the local-global adaptation~\cite{li2021benchmarking}, $1024^2$ input resolution exceeds the memory limitation of our GPU (24 GB), thus we choose a slightly smaller size $896^2$ during training. 

\textbf{For PVT (PVT-S) and Swin (Swin-T),} the input resolution is $1024^2$ during training. As these hierarchical ViTs already contain pyramid feature maps that are compatible with FPN, there is no need to apply further adaptations like ViT-B does.

\subsection{Semantic Segmentation on ADE20K}
ADE20K has 20K training images and 2K validation images of 150 categories. We use UperNet~\cite{xiao2018unified} following the code of \cite{bao2021beit}. The 16K-iteration polynomial learning rate schedule is applied with the first warm-up 1500 iterations. The AdamW optimizer is adopted with an initial learning rate of 1e-4, a weight decay of 0.05 and a batch size of 16 across 8 GPUs. The training resolution is $512^2$.

\textbf{For Vanilla ViT (ViT-B in this paper),} following MAE, the relative position bias is turned on during transfer learning, initialized as zero. The layer-wise lr decay~\cite{clark2020electra,bao2021beit} is set as 0.65 by default.

\textbf{For PVT (PVT-S) and Swin (Swin-T),} the layer-wise lr decay is not applied by simply setting it as 1.0, which always performs best in our experiments. 



\section{More Visualization Results}
We demonstrate more visualization results based on Swin-T under UM-MAE and SimMIM framework in Fig.~\ref{fig_sm_more_visual_cropped}. It is observed that the results of SimMIM are smoother than that of UM-MAE usually, yet are sometimes too smooth causing distortion. Although there are some differences in their visual reconstructions, the performances of fine-tuning are very close.

\end{document}

%% file: wrap/table_sm_ratio.tex
\begin{minipage}[ht]{0.38\textwidth}
\vspace{0pt}
    \renewcommand\arraystretch{1.2}
    \setlength{\tabcolsep}{6.pt}
    \footnotesize
    \centering
    \resizebox{\textwidth}{!}
    {
        \begin{tabular}{c|ccc}
        \hline
        SM Ratio & IN1K & ADE20K & COCO  \\ 

        \hline
          $20\%$ & 82.04 &  45.75 & {47.5}\\
          $25\%$ & 82.04 &\textbf{45.96} &  \textbf{47.7} \\
          $30\%$ & \textbf{82.10} &45.81 &  47.3 \\
        \hline
        \end{tabular}
    }
	\vspace{-4pt}
	\captionsetup{font={scriptsize}}
	\caption{\textbf{Secondary Masking Ratio.} Based on Swin-T, we pre-train models using different SM ratios for 200 epochs. The Top-1 Accuracy for IN1K, mIoU for ADE20K and AP for COCO are reported. $25\%$ performs good overall considering multiple tasks.
	}
\label{table_table_sm_ratio}
\vspace{-18pt}
\end{minipage}

%% file: wrap/figure_sm_ratio_pretrain_epoch.tex
\begin{minipage}[ht]{0.48\textwidth}
    \vspace{0pt}
	\begin{center}
		\setlength{\fboxrule}{0pt}
		\fbox{\includegraphics[width=0.96\textwidth]{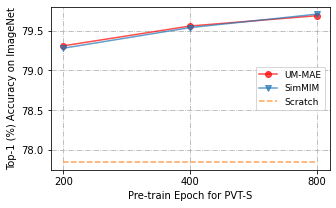}}
	\end{center}	
	\vspace{-16pt}
	\captionsetup{font={scriptsize}}
	\caption{\textbf{Pre-training schedules.} Longer pre-training epoch brings a slight improvement on fine-tuning accuracy. All the models (including the ``Scratch'') are fine-tuned/supervised for 100 epochs. 
	}
	\label{fig_pretrain_epoch}
	\vspace{-6pt}
\end{minipage}
\hspace{4pt}
\begin{minipage}[ht]{0.48\textwidth}
    \vspace{0pt}
	\begin{center}
		\setlength{\fboxrule}{0pt}
		\fbox{\includegraphics[width=0.96\textwidth]{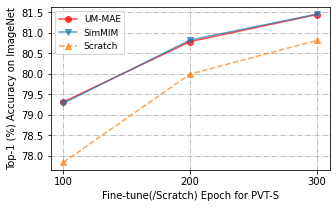}}
	\end{center}	
	\vspace{-16pt}
	\captionsetup{font={scriptsize}}
	\caption{\textbf{Fine-tuning schedules.} Longer fine-tuning epoch leads to consistent improvement, while the direct supervised scheme (i.e., Scratch) lags behind the self-supervised counterparts.
	}
	\label{fig_finetune_epoch}
	\vspace{-6pt}
\end{minipage}

%% file: wrap/table_large_model.tex
\begin{minipage}[ht]{0.4\textwidth}
\vspace{-10pt}
    \renewcommand\arraystretch{1.2}
    \setlength{\tabcolsep}{6.pt}
    \footnotesize
    \centering
    \resizebox{\textwidth}{!}
    {
        \begin{tabular}{l|c|c|c}
        \hline
        Method (Swin-L) & P-Size & EP-Size & IN1K \\\hline
        SimMIM~\cite{xie2021simmim}& $192^2$ & $192^2$ & 85.4 \\
        UM-MAE \textbf{(ours)}     & $256^2$ & $128^2$ & 85.3 \\
        \hline
        \end{tabular}
    }
	\vspace{-6pt}
	\captionsetup{font={scriptsize}}
	\caption{\textbf{IN1K Performance on large models.} The result of SimMIM is borrowed from the paper~\cite{xie2021simmim}. ``EP-Size'' denotes the Effective Pre-training Size. The reorganized compact 2D input in UM-MAE quarters the Pre-training Size from $256^2$ to $128^2$ by dropping $75\%$ tokens.
	}
\label{table_large_model}
\vspace{-20pt}
\end{minipage}

%% file: wrap/table_large_model_coco.tex
\begin{minipage}[ht]{0.49\textwidth}
\vspace{-10pt}
    \renewcommand\arraystretch{1.2}
    \setlength{\tabcolsep}{4.pt}
    \footnotesize
    \centering
    \resizebox{\textwidth}{!}
    {
        \begin{tabular}{l|c|c|cc}
        \hline
        HTC++ (Swin-L)  & Pre-train & Epoch & AP$^{\text{box}}$ &  AP$^{\text{mask}}$ \\\hline
        Baseline~\cite{liu2021swin}& IN22K, sup & 72  & 57.1 & 49.5 \\
        UM-MAE \textbf{(ours)}     & IN1K, unsup & 36 & \textbf{57.4} & \textbf{49.8} \\
        \hline
        \end{tabular}
    }
	\vspace{-6pt}
	\captionsetup{font={scriptsize}}
	\caption{\textbf{COCO Performance on large models.} The result of the Baseline refers to the paper~\cite{liu2021swin}. ``Epoch'': training schedule on COCO.
	}
\label{table_large_model_coco}
\vspace{-18pt}
\end{minipage}

%% file: wrap/table_discussion.tex
\begin{minipage}[ht]{0.46\textwidth}
	\vspace{0pt}
    \renewcommand\arraystretch{1.2}
    \setlength{\tabcolsep}{6.pt}
    \footnotesize
    \centering
    \resizebox{\textwidth}{!}
    {
        \begin{tabular}{lc|ccc}
        \hline
        Swin-T & Intermediate & mIoU & mAcc & aAcc  \\ 
        \hline
        Supervised & -- & 44.51 & 54.95  & 81.52 \\
        SimMIM & $\times$ & 40.25 & 50.89  & 80.10 \\
        UM-MAE & $\times$ & 41.14 & 52.13   & 80.01 \\
        \hline
        SimMIM & $\checkmark$ & 45.35 & 55.65   & 81.81 \\
        UM-MAE & $\checkmark$ & 45.96  & 56.55   & 81.58\\
        \hline
        \end{tabular}
    }
	\vspace{2pt}
	\captionsetup{font={scriptsize}}
    \caption{\textbf{The effectiveness of intermediate fine-tuning based on Swin-T when transferring it to downstream semantic segmentation tasks.} The intermediate fine-tuning is crucial for Pyramid-based ViTs self-supervised by various MIM frameworks. ``Intermediate'' denotes the usage of intermediate fine-tuning~\cite{bao2021beit,liu2021swinv2} on IN1K for 100 epochs, after the self-supervising process.
    }
    \label{table_discussion_intermediate}
	\vspace{-16pt}
\end{minipage}
\hspace{5pt}
\begin{minipage}[ht]{0.48\textwidth}
	\vspace{0pt}
	\newcommand{\specialcell}[2][c]{%
  \begin{tabular}[#1]{@{}c@{}}#2\end{tabular}}
    \renewcommand\arraystretch{1.2}
    \setlength{\tabcolsep}{5.pt}
    \footnotesize
    \centering
    \resizebox{\textwidth}{!}
    {
        \begin{tabular}{lcc|ccc}
        \hline
        Model & Pre-train & lw-lr decay & mIoU & mAcc & aAcc  \\ 
        \hline
        \multirow{3}{*}{ViT-B} & \multirow{3}{*}{\specialcell[t]{1600, \\ MAE}} & 1.00 & 45.87 & 56.19 & 82.46  \\
        & & 0.85 & 47.80 & 58.26 & \textbf{83.27} \\
        & & 0.65 & \textbf{48.15} & \textbf{58.99} & 83.05 \\
        \hline
        \multirow{3}{*}{Swin-T} &\multirow{3}{*}{\specialcell[t]{200, \\ UM-MAE}} & 1.00 & \textbf{45.96}  & 56.55  & \textbf{81.58} \\
        & & 0.85 & 45.80 & \textbf{56.68} & 81.45 \\
        & & 0.65 & 45.42 & 56.43 & 80.74\\
        \hline
        \end{tabular} 
    }
	\vspace{2pt}
	\captionsetup{font={scriptsize}}
    \caption{\textbf{The effectiveness of layer-wise learning rate decay (``lw-lr decay'' for short).} Under MIM, lw-lr decay is crucial for Vanilla ViT, but is harmful for Pyramid-based ViTs. ``lw-lr decay'' being $1.00$ means no decay applied. The pre-trained ViT-B model is downloaded from the official MAE github.
    }
    \label{table_discussion_lr_decay}
	\vspace{-16pt}
\end{minipage}